\documentclass[a4paper,fleqn]{cas-dc}

\usepackage[numbers]{natbib}

\usepackage{amssymb}
\usepackage{amsmath}
\usepackage{latexsym}
\usepackage{placeins}
\usepackage{xcolor}
\usepackage{subcaption}
\usepackage{babel}
\usepackage[utf8]{inputenc}
\usepackage{textcomp}
\usepackage{tabularx}

\def\tsc#1{\csdef{#1}{\textsc{\lowercase{#1}}\xspace}}
\tsc{WGM}
\tsc{QE}
\tsc{EP}
\tsc{PMS}
\tsc{BEC}
\tsc{DE}

\begin{document}
\let\WriteBookmarks\relax
\def\floatpagepagefraction{1}
\def\textpagefraction{.001}

\shorttitle{Style-Aware Facial Animation}

\shortauthors{Paier et~al.}

\title [mode = title]{Unsupervised Learning of Style-Aware Facial Animation from Real Acting Performances}                      

%
%
%
%
\author[1]{Wolfgang Paier}

\cormark[1]


\ead{wolfgang.paier@hhi.fraunhofer.de}




\author[1]{Anna Hilsmann}
\ead{anna.hilsmann@hhi.fraunhofer.de}

\author[1,2]{Peter Eisert}
\ead{peter.eisert@hhi.fraunhofer.de}


\affiliation[1]{organization={Fraunhofer Heinrich Hertz Institute},
    city={Berlin},
    country={Germany}}
    
\affiliation[2]{organization={Humboldt University},
    city={Berlin},
    country={Germany}}

%

\cortext[cor1]{Corresponding author}
\cortext[cor2]{Principal corresponding author}

%

\begin{abstract}
This paper presents a novel approach for text/speech-driven animation of a photo-realistic head model based on blend-shape geometry, dynamic textures, and neural rendering.
Training a VAE for geometry and texture yields a parametric model for accurate capturing and realistic synthesis of facial expressions from a latent feature vector.
Our animation method is based on a conditional CNN that transforms text or speech into a sequence of animation parameters.
In contrast to previous approaches, our animation model learns disentangling/synthesizing different acting-styles in an unsupervised manner, requiring only phonetic labels that describe the content of training sequences.
For realistic real-time rendering, we train a U-Net that refines rasterization-based renderings by computing improved pixel colors and a foreground matte. We compare our framework qualitatively/quantitatively against recent methods for head modeling as well as facial animation and evaluate the perceived rendering/animation quality in a user-study, which indicates large improvements compared to state-of-the-art approaches.

\end{abstract}




\maketitle
\sloppy
\section{Introduction}
Modeling and animation of virtual humans play an essential role in many applications such as virtual reality (VR), video games, or movie productions.
Especially, the synthesis of realistic talking head videos from speech data received great attention, as this will be an important feature in future applications like virtual assistants or service agents.
While recent approaches have made great progress in terms of realism and animation quality, automatically synthesizing believable facial animations from speech or text still remains a challenge.

We present a new approach for creating an animatable and photo-realistic 3D head model from multi-view video footage of a real actor, together with a neural animation model based on emotional performances. The approach allows for synthesizing new emotional speech sequences from text or speech and manipulates speaking styles as well as emotions with a low dimensional latent style vector.
Our head model is based on a hybrid representation that combines 3D mesh-based geometry, dynamic textures, and neural rendering.
A personalized statistical geometry model captures rigid motion as well as large-scale deformations.
Additionally, we extract dynamic textures from multi-view images in order to capture facial details, fine motion, and the appearance of complex regions (e.g.~mouth cavity, eyes, or hair), which cannot be represented by the underlying mesh alone.
Based on mesh and dynamic texture data, we create a parametric face model by training a joint variational autoencoder (VAE) for face texture and geometry.
This parametric representation gives control over displayed facial expressions and allows for animation from text or speech.
While this is a convenient and efficient way to represent facial expressions in every detail, it is insufficient for photo-realistic rendering that requires highly detailed geometry even in complex regions (e.g.~mouth, eyes, hair) and accurate material and reflectance information.
To account for this, we employ a pixel-to-pixel translation network that refines rasterization-based head images such that the appearance of hair, silhouette and other view-dependent effects are correctly reproduced.
We train the render network in a self-supervised manner, which minimizes the pre-processing overhead for multi-view data since foreground/background segmentation is learned automatically during training.

For synthesis and animation, we train our neural animation model based on automatically generated phonetic annotations, which represent the speech content in all captured video sequences.
Using a pre-defined mapping, we convert phonetic annotations to visemes IDs, which serve as input features.
We chose to rely on visemes, because of their speaker independence and versatility (e.g. both speech, as well as text, can be easily converted to viseme sequences).
As we aim at synthesizing facial performances similar to that of a real actress/actor, we design our network architecture in a way that allows for unsupervised learning of emotions, natural variations of facial expressions, and head movements during speech.
This has several advantages: first, no additional manual effort for annotating emotions or speaking style is required; second, we do not need to capture an exhaustive training database that contains all combinations of speech sequences, emotions, and emotion intensities; third, our actress can focus on an authentic performance rather than acting according to a target emotion, which increases the naturalness of the training data, and thus also the realism and authenticity of the synthesized animations.
After training, we can synthesize realistic talking-head videos from text or speech while being able to manipulate speaking style and emotions with a low-dimensional control vector.

\section{Related Work}
\label{sec:relatedwork}

\subsection*{Face Capture and Modeling}
Creating controllable models of human heads and faces has been extensively studied in the field of computer vision and computer graphics.
One of the most popular approaches are morphable models \cite{Blanz1999,Egger2020}.
In order to deal with their limitations, researchers proposed different strategies to increase the quality of captured facial expressions~\cite{Cootes98, eigi98, Blanz1999, Vlasic2005, Cao2014} 
such as the personalization of blend-shapes \cite{Weise2009,Li2010,Weise2011,Bouaziz2013,Garrido2013,Li2013}, computation of corrective shapes during run-time \cite{Li2013}, or the extension of linear models to capture 
multiple additional attributes such as identity, texture, and light \cite{Thies2015}.
However, one major drawback of linear models is the lack of expressive power to correctly represent complex areas like the oral cavity, eyes, or hair.
As a result, purely model-based approaches employ either 'hand crafted' solutions (e.g.~oral cavity) or simply ignore them~\cite{Garrido2013, Thies2015}.
Alternatively, hybrid or image-based methods solve this problem by representing facial performances in geometry and texture space~\cite{Paier2017, Casas2014, Dale2011, Lipski11, Kilner06, Borshukov2006, Carranza2003a}.

With the advent of deep learning, more sophisticated approaches have been proposed~\cite{Tewari2017, Tewari2018, Tewari2019, Chai2020, Mallikarjun2021}.
For example, Tewari et al.~\cite{Tewari2017} and Chai et al.~\cite{Chai2020} use large collections of unlabelled face images and video in order to train and refine existing linear face models in a self-supervised or unsupervised manner.
In order to improve the rather simple image formation model of previous approaches, Dib et al.~\cite{Dib2021} employ a differentiable ray tracer, which allows for more advanced materials, better illumination, and self-shadowing.

Siarohin et al.~\cite{Siarohin2019} proposed an approach that directly animates an image of a person using 2D affine transformations, which further reduces the effort of creating a controllable face model.
While their system works well in most cases, large changes in the head pose typically create unnatural distortions.
Ren et al.~\cite{Ren2021} try to avoid this limitation by employing a 3D morphable head model that allows for larger changes of the head pose.
Wang et al.~\cite{Wang2021} improve the quality of animated portraits by explicitly predicting features for appearance, expression, head pose, and canonical 3D key points to guide the image formation process, whereas 
Hong et al.~\cite{Hong2022} propose a depth-aware generative adversarial network (GAN) that explicitly learns predicting depth values for each pixel in order to improve the quality of the warped face image. 

While these approaches have the advantage that controllable face models for a specific person can be generated with low effort, they usually do not yield highly detailed and realistic 3D head representations.
In order to solve this task, deep generative face models have been introduced \cite{Lombardi2018, Li2020, Chandran2020, Bi2021, nagano2018, Slossberg2018}.
For example, Lombardi et al.~\cite{Lombardi2018} capture multi-view facial video with 40 synchronized machine vision cameras and 200 LED point lights to promote uniform illumination.
Using a highly detailed blend shape model, they recover face geometry and view-dependent textures per frame, which are then used for training a variational autoencoder (VAE)~\cite{kingma2013autoencoding}.
After training, the decoder can be used as a parametric non-linear face model that reconstructs dynamic geometry and expression- as well as view-dependent textures.
Li et al.~\cite{Li2020}as well as Chandran et al.~\cite{Chandran2020} propose deep neural face models, which are capable of representing facial expressions as well as identities.
Chandran et al.~train their neural face model with registered and textured face meshes of 224 subjects showing 24 predefined expressions.
Two separate VAEs are used to extract identity and expression information.
A joined decoder uses a latent identity and expression vector to reconstruct the target geometry as well as a low-resolution albedo map.
Using a super-resolution network, they compute the final high-resolution albedo map.
Li et al.~\cite{Li2020} build their neural face model using two generative adversarial networks (GAN) based on the styleGAN architecture.
Ma et al.~\cite{Ma2021} proposed a codec avatar that allows for efficient rendering of detailed head avatars even on mobile devices.
More recently, Grassal et al.~\cite{Grassal2022} proposed an approach for generating neural head models from monocular RGB video that are able to dynamically refine geometry and texture, which allows for modeling complex geometry that cannot be captured by the underlying morphable model.

\subsection*{Neural Rendering}
In recent years, novel scene representations~\cite{Mildenhall2019, Sitzmann2019, Lombardi2019b,Pumarola2020, Mildenhall2020,Tewari2020STAR,Yu2021, Reiser2021, Hedman2021,Gafni2021} and sophisticated methods for image synthesis based on neural networks~\cite{Kim2018,MartinBrualla2018,Wang2018,Thies2019,Aliev2020} have been proposed that allow for creating truly photo-realistic renderings of humans.
For example, Martin-Buralla et al.~\cite{MartinBrualla2018} use a deep convolutional network to refine the renderings of 3D point clouds in virtual reality applications.
Wang et al.~\cite{Wang2018} implemented a video-to-video translation network that synthesizes realistic videos of arbitrary objects or scenes based on semantic segmentation videos,
whereas Thies et al.~\cite{Thies2019} propose a deferred neural rendering approach to create photo-realistic renderings of 3D computer graphic models.
Kim et al.~\cite{Kim2018} and Prokudin et al.~\cite{Prokudin2020} combine neural rendering with parametric models of humans and human heads, which allows for photo-realistic rendering of animatable CG models.
Recently, novel implicit scene representations based on neural networks have been proposed \cite{Mildenhall2019, Sitzmann2019, Lombardi2019b, Pumarola2020, Mildenhall2020,Tewari2020STAR}.
The core idea is to capture the 3D structure and appearance of an object/scene with a non-linear function that depends on the 3D position in space as well as viewing direction and predicts color as well as volume density.
While the key idea is simple, it enables the network to represent fine structures such as hair fibers, objects with complex reflective properties such as glass and metal but also dynamic effects such as smoke.
Although they have not been specifically designed for representing human faces, they provide useful capabilities that classic computer graphic models lack and that can help synthesizing more realistic faces/heads (e.g.~rendering hair and complex geometries). A big drawback, however, is the high computational complexity \cite{Yu2021, Reiser2021, Hedman2021,Gafni2021}.
While M\"uller et al.~\cite{Mueller2022} propose a new approach that supports very fast training as well as real-time rendering, they still lack semantic control, which is essential for animation.

\subsubsection*{Neural Animation}
With the advances in the processing of natural language and sequential data, several new methods for facial animation from speech have been proposed~\cite{Suwajanakorn2017,Zhou2018,Fried2019,Thies2020,Prajwal2020,Zhang2021,Guo2021,Zhou2021}.
For example, Suwajanakorn et al.~\cite{Suwajanakorn2017} train a recurrent neural network (LSTM) on a large collection of weekly presidential addresses (17 hours in total) to predict mouth landmarks according to a speech signal.
Based on these mouth landmarks, they generate a realistic mouth texture, which is integrated into a re-timed target video.
Zhou et al.~\cite{Zhou2018} tackle the problem of animating a production face rig from a speech signal based on a small amount of training data using transfer learning and complementary objective function for which enough data is available.
Other researchers published similar methods~\cite{Thies2020,Zhang2021,Guo2021}: Thies et al.~\cite{Thies2020} use a CNN-based approach together with neural rendering to synthesize photo-realistic videos from speech, Zhang et al.~\cite{Zhang2021} use a GAN-based approach that could synthesize plausible eye blinks and Guo et al.~\cite{Guo2021} introduced a new NeRF-based approach for talking head synthesis from speech.

With the availability of larger audio-visual datasets, several methods for speaker-aware facial animation~\cite{Cudeiro2019,Chen2020,Zhou2020} have been proposed: Cudeiro et al.~\cite{Cudeiro2019} use one-hot-encoding to model subject-specific speaking styles, Zhou et al.~\cite{Zhou2020} extract a speaker identity vector from the speech signal that helps synthesizing speaker dependent face landmark displacements, but this does not allow for representing emotion or manipulating the speaking style of the animated character.
Chen et al.~\cite{Chen2020} propose a method that synthesizes more natural head motion by explicitly representing it in 3D space.

Recent works started modeling the effects of emotion and speaking styles~\cite{Karras2017,Wang2020,Eskimez2020,Cheng2021,Ji2021}:
for example, Karras et al.~\cite{Karras2017} propose a system that predicts vertex offsets according to an input speech signal.
In addition, they learn a specific style vector for each training sample in order to capture natural variations of facial expressions, however, their style features are difficult to interpret and their approach does not model important features such as eyes, oral cavity, texture, or head motion.
Most closely related to our method are Eskimez et al.~\cite{Eskimez2020}, Cheng et al.~\cite{Cheng2021}, and Ji et al.~\cite{Ji2021}.
However, their models cannot be trained without an exhaustive (emotion) database or annotated emotions.
Eskimez et al.~\cite{Eskimez2020} as well as Cheng et al.~\cite{Cheng2021} require a content signal~(i.e. speech/text) as well as pre-defined emotion labels.
While the former are able to synthesize a complete face video based on the emotion condition,  Cheng et al.~\cite{Cheng2021} use this information only for the synthesis of eye expressions.
Ji et al.~\cite{Ji2021} do not directly predict from emotion labels, but they require pairs of training sequences with the same content and different emotion in order to employ a cross-reconstruction training.
Although this reduces the demands on the training data, it is not enough to facilitate training with regular video sequences that do not contain emotion labels and are not captured multiple times with different emotions.
In contrast, our approach does neither require emotion labels nor multiple versions of each training sequence in order to disentangle animation style and content.
Moreover, our method is able to synthesize realistic expression and motion parameters conditioned on automatically learned style features for mouth, eyes, and head pose.


\subsubsection*{Contributions}
\textbf{Hybrid 3D Head Model:}
We present an approach for creating a personalized and animatable 3D head avatar from video data of a real person.
Similar to other methods~\cite{Kim2018,Thies2019,Zhou2020,Gafni2021,Ren2021,Grassal2022}, we use a statistical geometry model to represent rigid motion and large-scale deformations of the face,
but instead of relying on the existing expression space of the underlying geometry model, we compute a refined expression space that allows representing and synthesizing all facial expressions of the captured person.
This is possible because the personalized dynamic texture model captures complex areas such as the eyes or the oral cavity (even if not modeled in geometry) as well as fine details and small deformations
that cannot be represented by the geometry model alone.
Another similar method was proposed by Ma et al.~\cite{Ma2021}. Starting from low-resolution mesh, they synthesize a dense mesh that refines inaccurate/missing parts~(e.g. tongue) of the coarse underlying tracking mesh.
However, this requires high-resolution multi-view stereo matching for the training videos, while our approach can essentially be used without detailed 3D reconstruction of the captured scenes.

\textbf{Self-Supervised Neural Rendering}:
In contrast to previous approaches~\cite{Kim2018,MartinBrualla2018,Aliev2020,Prokudin2020,Cheng2021,Zhang2021}, we  train the rendering model in a self-supervised fashion.
This allows for synthesizing photo-realistic head images in real-time together with alpha masks that separate foreground from background, which minimizes pre-processing overhead and allows inserting the rendered head image into new backgrounds.
While the approach of Gafni et al.~\cite{Gafni2021} offers similar capabilities, we can show that our system yields higher image quality and clearer alpha masks.

\textbf{Style-Aware Neural Animation}:
In contrast to existing approaches~\cite{Suwajanakorn2017,Chen2020,Eskimez2020,Thies2020,Wang2020,Guo2021,Cheng2021,Zhang2021,Ji2021}, our system can synthesize authentic speech performances after being trained on a challenging dataset with a low number of training samples, unannotated and varying emotions/speaking styles as well as strong head motion.
\begin{figure*}[h!]
\centering
\includegraphics[width=0.9\textwidth]{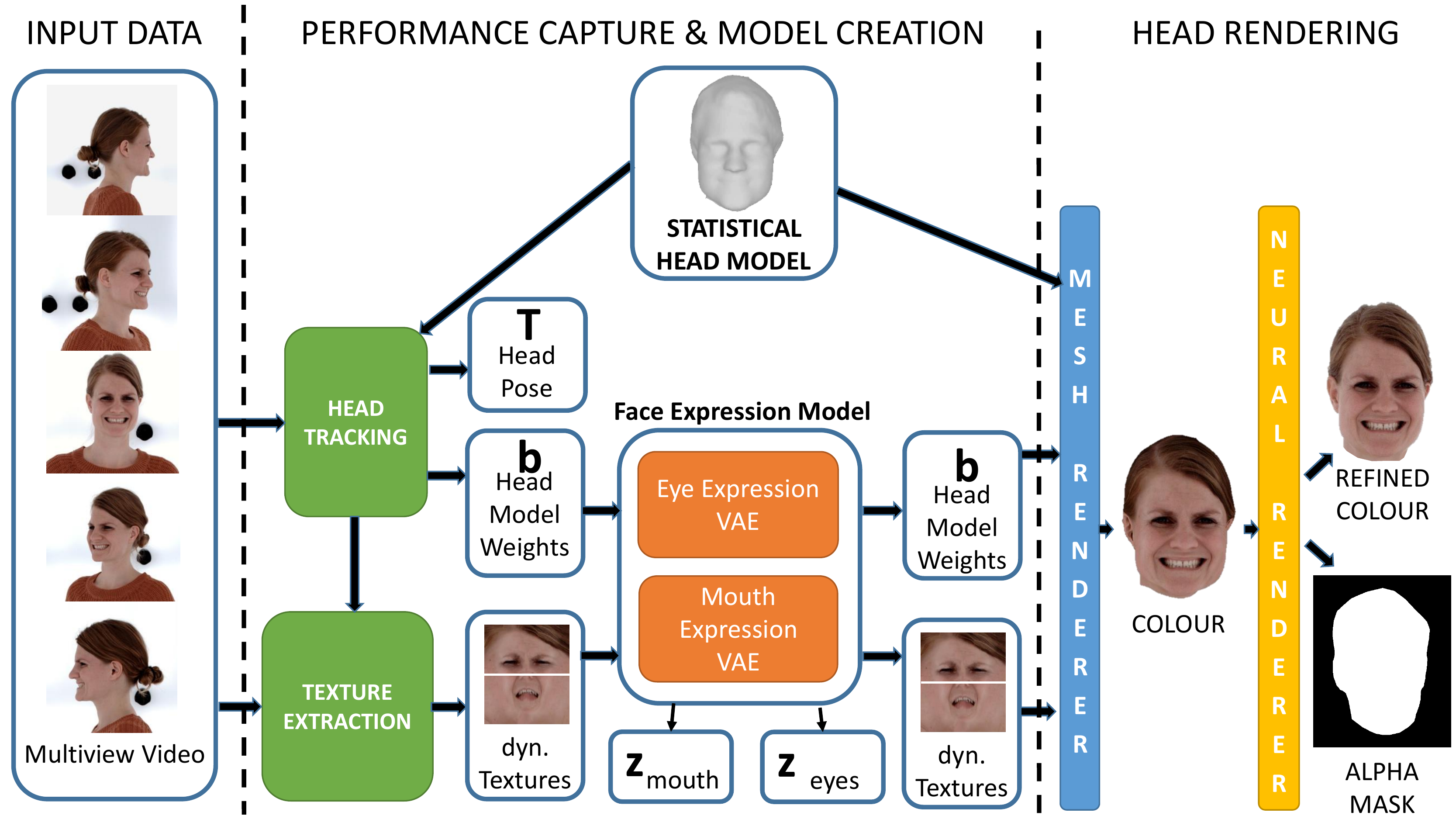}
\protect\caption{Illustration of the overall architecture of the proposed capture, model creation, and rendering approach. A hybrid performance capture pipeline (green) computes dynamic head geometry and textures from multi-view video.
The extracted geometry-plus-texture sequences are used to train a deep generative head model (VAE), which allows for synthesis, animation, and editing. Finally, a neural re-rendering network creates realistic images of the 3D head model.}
\label{fig:overall}
\end{figure*}
\FloatBarrier
This is beneficial as actors can perform more authentically since strong facial expressions and head motions do not degrade the quality of synthesized talking-head videos, but actually increase realism.
Training our animation network in a semi-supervised way (i.e.~with phonetic annotations via forced alignment) allows for automatically discovering speaking styles and emotions that occur in natural acting performances.
Moreover, we show a simple but effective way of \textbf{visualizing the learned style/emotion space}, which enables users to shape synthesized animations according to their needs.

\textbf{Extensive Evaluation and User Study:}
We evaluate our system qualitatively as well as quantitatively against several recent approaches for modeling, animation, and rendering.
Since traditional error metrics do not always capture the perceived quality of rendered head videos, we conduct an extensive user study in order to evaluate the perceived rendering quality, the audio-visual synchronization as well as the realism of synthesized facial expressions and head motion.
\begin{figure*}[t]
\includegraphics[trim=0mm 10mm 0mm 0mm,clip,width=\textwidth]{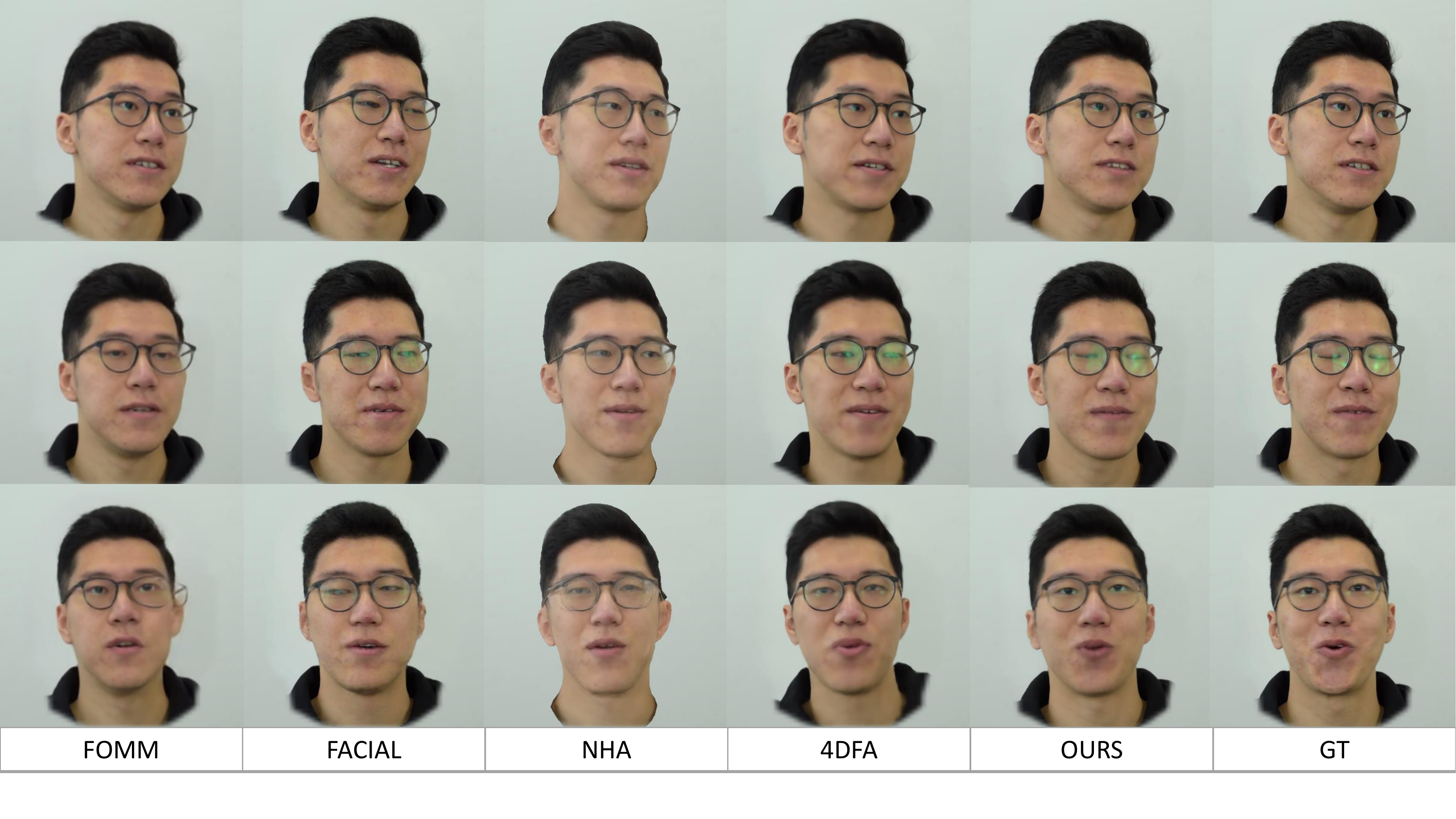}
\protect\caption{Comparison of the rendering quality of different methods. Each row corresponds to a certain time frame in the validation set, where each column shows the rendering result obtained with a certain approach. From left to right we show rendered heads using the method of Siarohin et al.~\cite{Siarohin2019} (FOMM), Zhang et al.~\cite{Zhang2021} (FACIAL), Grassal et al.~\cite{Grassal2022} (NHA), Gafni et al.~\cite{Gafni2021} (4DFA), our proposed approach and the ground truth. Especially mouth~(third row), eyes~(second row) and glasses~(third row) can deviate strongly from the ground-truth.}
\label{fig:render-results1}
\end{figure*}
\section{System Overview}
\label{sec:overview}
In a first step, we create a personalized parametric head model that represents 3D geometry, head motion, facial expression, and appearance of the captured person~(Sec.~\ref{sec:capture}). 
The model is learned from multi-view data of an actor captured with a multi-view video rig and a microphone for the speech signal.
For animation control, we use a latent parameter vector of an autoencoder, which essentially stores the captured facial performance as a sequence of low-dimensional parameter vectors.
This latent vector can be driven from text, speech, or video input data, exploiting a corresponding mapping network.

Using forced alignment, we automatically annotate the captured voice signal with time-aligned phonemes in order to represent the speech content.
We train a neural animation model together with a speech style encoder from annotated speech signals and corresponding animation parameters, which allows for disentangling the style and content of the captured performance~(Sec.~\ref{sec:animation}).
After training, we can condition our neural animation model with a low-dimensional style/emotion vector that allows for synthesizing authentic and emotional talking head videos.

To increase realism, we train a neural rendering model that refines rasterization-based images of our head model, which allows for synthesizing photo-realistic videos of the animated head model~(Sec.~\ref{sec:rendering}).
\section{Head Model Creation and Rendering}
\label{sec:capture}
\begin{figure*}[t]
\includegraphics[trim=0mm 10mm 0mm 0mm,clip,width=\textwidth]{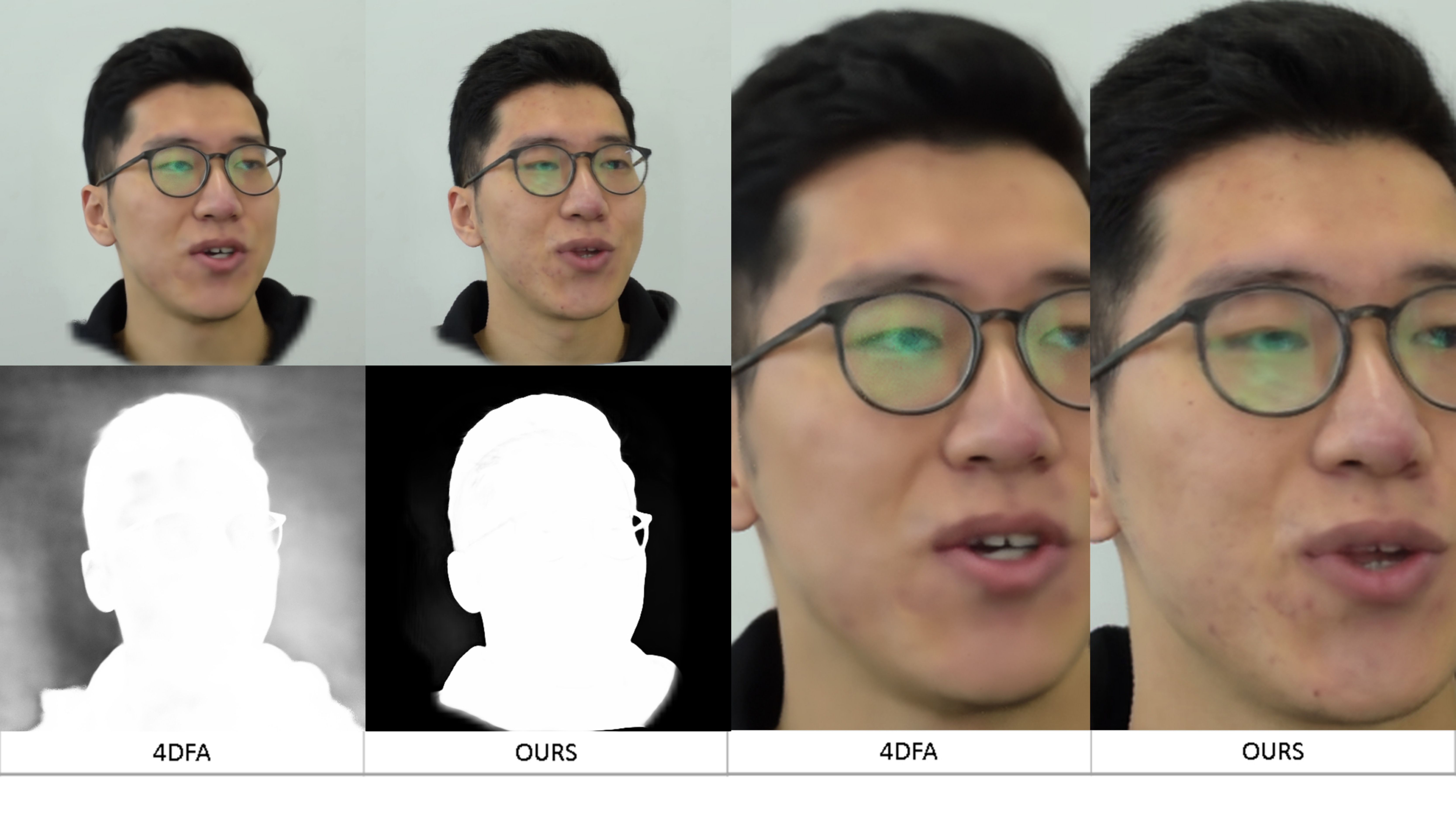}
\protect\caption{This figure gives a visual comparison between the results of 4DFA and our approach.
The left side shows rendered images and foreground masks, while the right side shows a zoom-in on the face. Our approach is able to produce more accurate foreground masks as well as more details on skin, eyebrows, and hair.}
\label{fig:render-results3}
\end{figure*}
The core of our system is constituted by a photo-realistic animatable 3D head model that is computed from multi-view video footage to ensure that it
perfectly resembles the appearance of the captured person.
Our model creation process is based on \cite{paier2020} and consists of three stages that are illustrated in figure~\ref{fig:overall}.
First, we employ a statistical head model to recover pose and approximate head geometry for each captured frame using automatically detected landmarks~\cite{Kazemi2014} and optical flow.
Statistical models are sufficient to capture large-scale variations with only a small number of parameters, but they typically fail to reproduce fine details, small motions, and complex geometry deformations (e.g.~in the oral cavity).
To account for this, in a second step, we extract dynamic head textures in addition to the approximate geometry assuming that the head model already possesses valid texture coordinates.
This allows for capturing missing details, small motions, and appearance changes in texture space.
While the original approach computes the texture sequence by solving a discrete optimization task over all frames,
we improve computational efficiency by solving the objective for a smaller set of keyframes and approximating the solution for other frames via the nearest keyframe.
This allows for processing longer sequences, reduces computation time, and introduces implicit temporal smoothness.
After geometry recovery and texture extraction, each captured performance is represented by a sequence of textured meshes with corresponding geometry.
More precisely, each frame is represented by rigid motion parameters $\mathbf{T}$, blend-shape weights $\mathbf{b}$, and an RGB image as texture.
Based on this representation, we train a deep generative face model that reconstructs blend-shape weights $\mathbf{b}$ as well as face textures from a shared low-dimensional expression vector, thereby enabling natural, plausible, and realistic facial animation. 
The joined parametric representation ensures that texture and geometry always fit together such that facial expressions are reconstructed correctly.
We create two local animation models that represent mouth and eyes by manually selecting two rectangular regions of interest in texture space.
Since pre-computed texture coordinates are provided by the underlying head template, this step has to be performed only once as the selected mouth and eye regions can be re-used for the creation of further head avatars.
For the model creation, a variational autoencoder~\cite{kingma2013autoencoding} architecture is employed.
The encoder receives shape weights $\mathbf{b}$ as well as a cropped texture region from the performance capture stage as input and transforms them into a low dimensional latent representation $\mathbf{z}$ of the facial expression.
The decoder receives the noisy code vector $\mathbf{z}$ and reconstructs shape weights $\mathbf{b}$ as well as the texture for mouth/eyes.

In order to improve texture reconstruction quality, we employ an adversarial training strategy using a patch-based discriminator network~\cite{Isola2016} that is trained simultaneously with the autoencoder.
During training, we minimize the $L_1$ texture reconstruction loss, the adversarial texture loss, and the $MSE$ between predicted and target shape weights.
For better control over facial expressions, we create two expression models, one for the mouth and one for the eyes.
After training, all captured multi-view sequences are represented with a single parameter vector consisting of 3D head pose $\mathbf{T}$ as well as expression vectors $\mathbf{z}_{mouth}$ and $\mathbf{z}_{eyes}$ for the corresponding face regions.
The separate representation of mouth and eyes increases the flexibility as both regions can be controlled independently of each other.
Moreover, it enables the use of a specialized animation prior for the eyes, which further increases the realism synthesized talking-head-video (section \ref{sec:animation-prior}).
\subsection{Neural Rendering}
\label{sec:rendering}
\begin{figure*}[ht!]
\includegraphics[trim=0mm 0mm 0mm 0mm,clip,width=\textwidth]{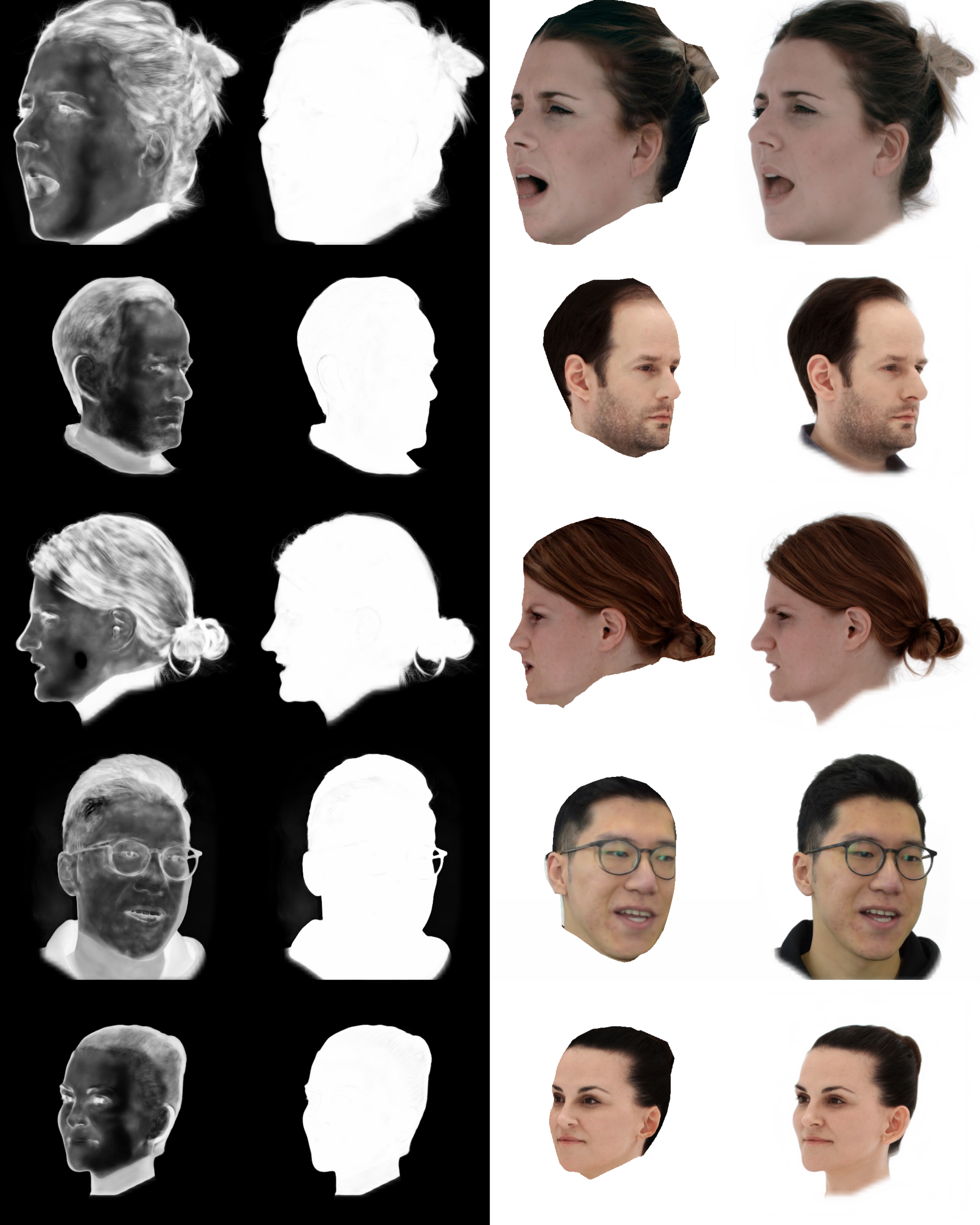}
\protect\caption{This figure contains rendering results based on different datasets.
The leftmost column shows the refinement weights $\beta$ followed by the foreground mask $\mathcal{F}$, the initial mesh-based rendering, and the final output $\mathcal{I}_{out}$. High intensity in the refinement mask indicates strong corrections (e.g.~neck, hair, sometimes mouth), while low intensity indicates that the mesh-based rendering provides already correct pixel colors.}
\label{fig:render-input-output}
\end{figure*}

While the traditional mesh-plus-texture representation correctly reconstructs the facial performance as well as most of the appearance, some important details cannot be reproduced, see figure \ref{fig:render-input-output}.
For example, due to the rough geometry model, the silhouette appears smooth and unnaturally clean.
Looking at the model from the side while the mouth is open reveals that the underlying statistical shape model does not properly capture the oral cavity. 
Moreover, the hair is not reconstructed faithfully as view-dependent appearance effects, and fine hair strands near the silhouette or eyelashes are missing.
Some of these shortcomings could be resolved by increasing the complexity of the geometry model or introducing specialized part models for the oral cavity, eyes \cite{Thies2018}, or hair \cite{Hu2017}.
However, this would also require a more complex capture setup and could make manual intervention necessary while still not achieving the desired level of realism.

Instead, we employ a self-supervised approach based on pixel-to-pixel translation, where our render network receives the mesh-based rendering as input and predicts a refined head image as well as weight masks that help separating foreground from background.
This simplifies the training process as we do not need to pre-compute foreground masks because providing clean plate images (as background models) for each camera is sufficient.
The image formation model (\ref{eq:imageformation}) of our neural rendering approach can be expressed as a convex combination of the mesh-based rendering $\mathcal{I}_{orig}$, the corrective image $\mathcal{I}_{corr}$ and the static background model $\mathcal{I}_{backg}$, where each image contributes according to spatially varying weight maps $\alpha$, $\beta$, and $\gamma$.

\begin{equation}
\begin{split}
    \mathcal{I}_{out} = \alpha\mathcal{I}_{orig} + \beta\mathcal{I}_{corr} + \gamma\mathcal{I}_{backg} \\
    \alpha + \beta + \gamma = 1
\end{split}
\label{eq:imageformation}
\end{equation}
During experiments, we found the explicit weighting of the corrective image advantageous as it allows the network to disable refinement in certain areas (e.g.~background or areas that are already well represented by the textured mesh).

\begin{figure*}[ht!]
\includegraphics[trim=0mm 30mm 0mm 0mm,clip,width=\textwidth]{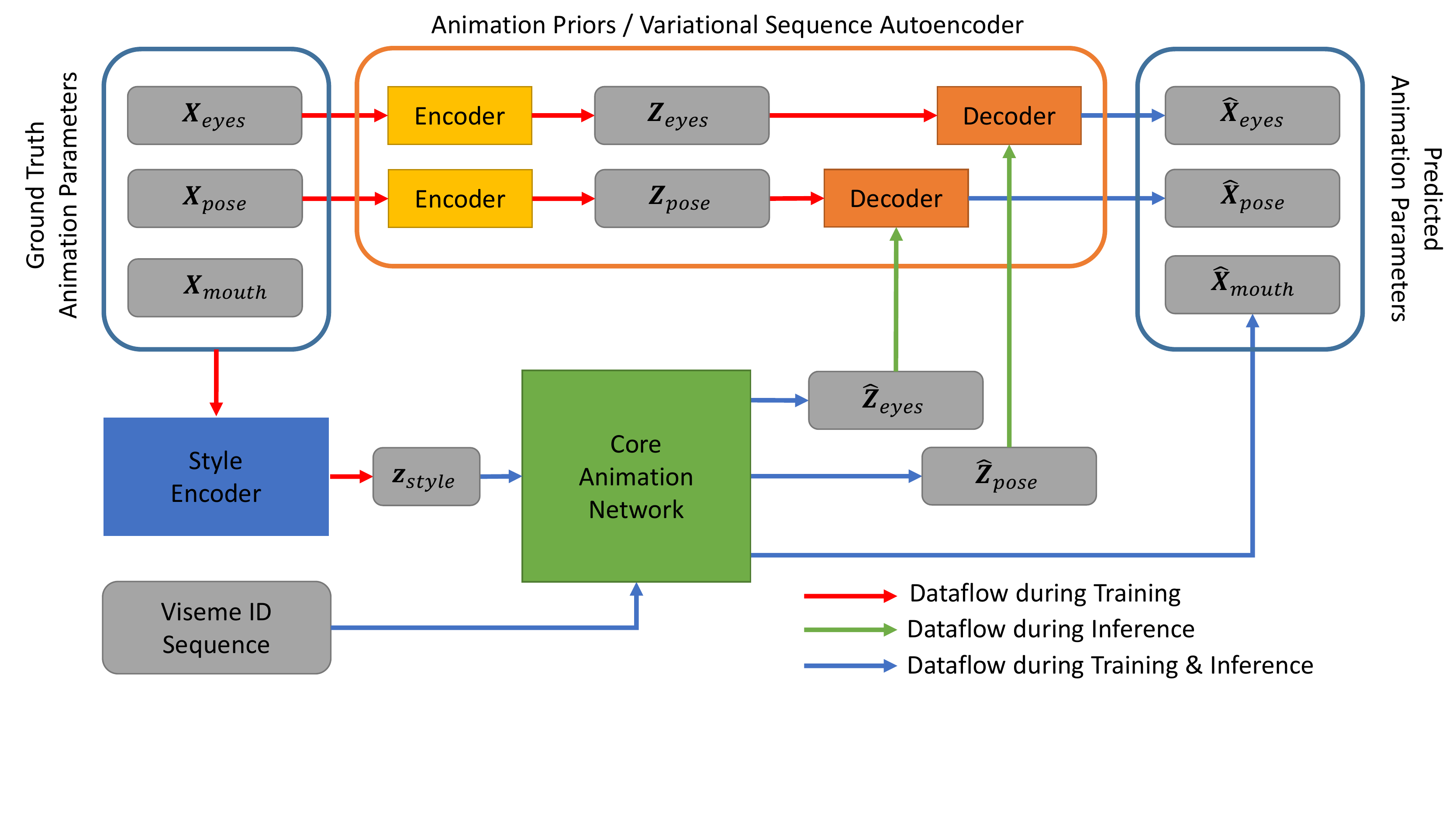}
\protect\caption{Illustration of the high-level animation architecture.
Two animation priors, one for the eyes and one for the head pose are trained as variational sequence auto-encoder that reconstruct the ground-truth animation parameters from a compressed latent sequence $\mathbf{Z}_{eyes}$ and $\mathbf{Z}_{pose}$.
During training, the style encoder extracts a latent style feature $\mathbf{z}_{style}$ from the ground-truth parameters.
Based on the style-feature and the input viseme IDs, the animation network predicts mouth animation parameters $\mathbf{\hat{X}}_{mouth}$ as well as the latent sequence parameters $\hat{\mathbf{Z}}_{eyes}$ and $\hat{\mathbf{Z}}_{pose}$, which are converted into eye and pose animation parameters ($\mathbf{\hat{X}}_{eyes}$ and $\mathbf{\hat{X}}_{pose}$) by the corresponding sequence decoder.}
\label{fig:anim-arch}
\end{figure*}
Especially when the static background model is not sufficiently accurate, a render model without explicit weighting of $\mathcal{I}_{corr}$ would try refining the background as well, which would result in the wrong foreground ($\alpha+\beta$) and background ($\gamma$) weights.
Our re-rendering network is based on a U-Net architecture \cite{Ronneberger2015} and the input tensor contains the RGB colors of the mesh-based rendering.
The output tensor consists of six channels: RGB color plus three channels that contain the weight maps $\alpha$, $\beta$, and $\gamma$.
We use a U-Net with five layers, 64 filters (instead of 32) at the first layer and a scaling factor of 4 (instead of 2) at each layer as this yields higher rendering quality while still achieving high rendering speed (12 milliseconds per frame).

Training our network requires the captured frame, the initial rendering $\mathcal{I}_{orig}$ of the textured head model, an empty background frame $\mathcal{I}_{backg}$ as well as a binary foreground/background mask of the initial rendering.
The corrective image $\mathcal{I}_{corr}$ is predicted by the first three channels of the output tensor using a $tanh$ activation.
The weight maps $\alpha$, $\beta$, and $\gamma$ are predicted by the last three channels of the output tensor using a soft-max activation.

The image reconstruction is guided by the VGG loss \cite{Johnson2016} between the reconstructed frame $\mathcal{I}_{out}$ and the captured frame.
Moreover, we minimize an adversarial loss $\mathcal{E}_{adv}$ by introducing a patch-based discriminator \cite{Isola2016} that aims at distinguishing between real and synthesized images on a patch basis.
The full objective function (\ref{eq:fullobjective}) is given by

    \begin{equation}
    \mathcal{E} = w_{vgg}\mathcal{E}_{vgg} + w_{adv}\mathcal{E}_{adv} + w_{pri}\mathcal{E}_{pri} + w_{bin}\mathcal{E}_{bin} + w_{reg}\mathcal{E}_{reg}, 
    \label{eq:fullobjective}
    \end{equation}

which is minimized using the Adam optimizer with default parameters, an initial learning rate of 0.0001, exponential learning rate scheduling (0.95), and a batch size of four.
During training, we randomly scale, rotate, and translate source and target images in order to support rendering at different distances (scales) or partially visible heads.
In order to minimize flickering in $\mathcal{I}_{out}$, we smooth the mesh-based rendering at the border between the head and background using a Gaussian filter.
The render network for each subject is trained for thirty epochs, which corresponds to approximately 93k iterations.
The weights for each loss in (\ref{eq:fullobjective}) are: $w_{vgg}=1$, $w_{adv}=0.1$, $w_{pri}=0.1$, $w_{bin}=0.1$, $w_{reg}=0.001$.
The first 5000 iterations are considered as a warm-up phase where $w_{bin}$ and $w_{adv}$ are set to zero.
Starting from iteration 5000, the weights of $w_{bin}$ and $w_{adv}$ are linearly ramped up such that they reach their target value after 1000 steps at iteration 6000.

\subsubsection{Mask Regularization}
\begin{figure}[t]
\centering
\includegraphics[trim=0mm 10mm 0mm 0mm,clip,width=1.0\linewidth]{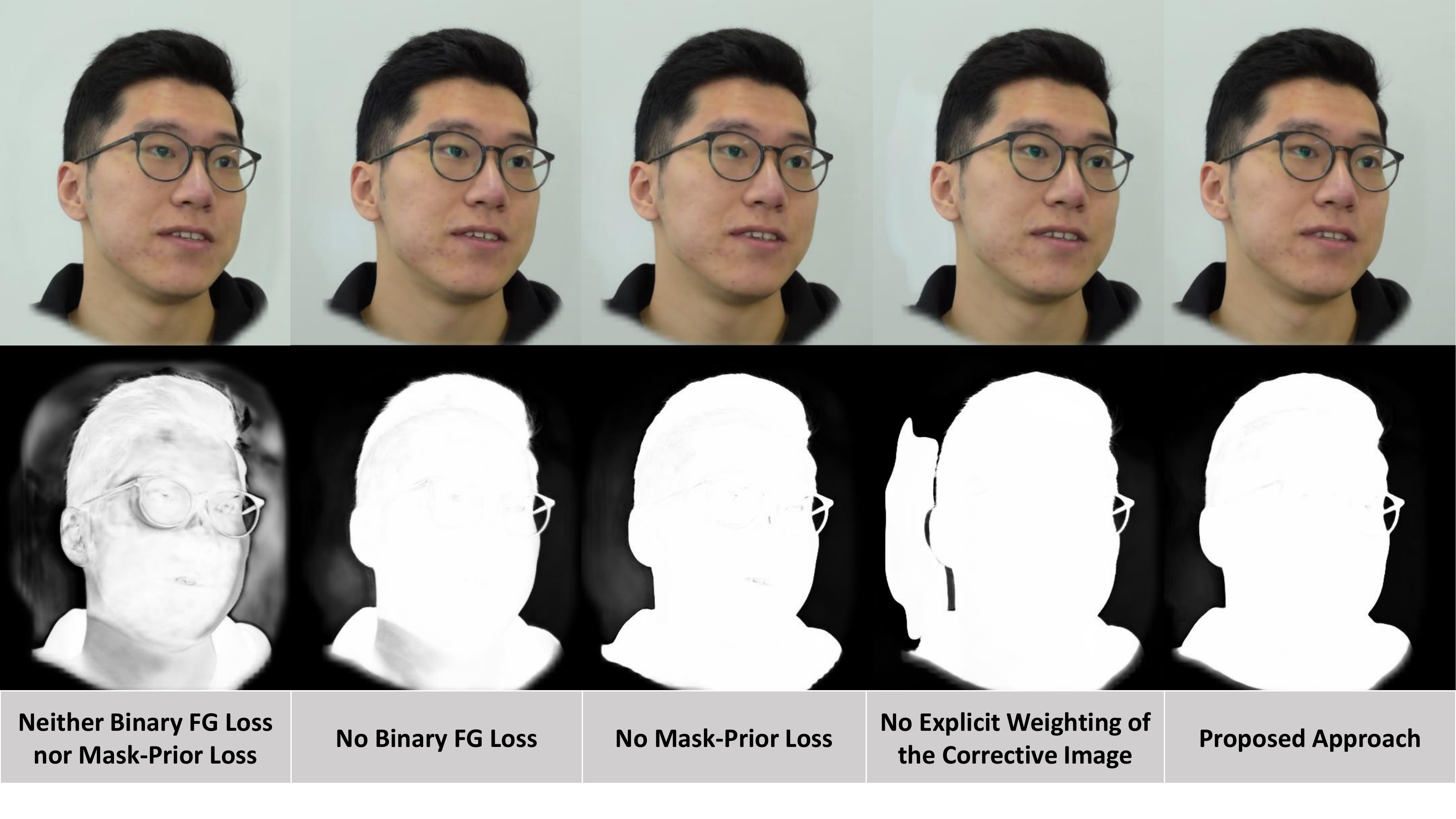}
\protect\caption{This figure illustrates how the proposed image formation model and mask regularization affect the synthesized foreground masks.
While the rendered images look very similar, the quality of foreground masks is strongly affected.}
\label{fig:mask-ablation}
\end{figure}

In order to better constrain the refinement process, we implement two additional loss functions to create more accurate and clean foreground masks $\mathcal{F}$.
This is especially important when the background model is not accurate (e.g.~due to dynamic shadows) since the render network might learn to refine the background as well (figure \ref{fig:mask-ablation}, leftmost image).
    \begin{equation}
    \mathcal{F} = \alpha + \beta
    \label{eq:foregroundmask}
    \end{equation}
First, we generate a dilated as well as an eroded version of the initial binary render mask $\mathcal{M}$.
Outside the dilated mask $\mathcal{M}_{d}$, the foreground likelihood should be zero (right side of (\ref{eq:maskprior})), while inside of the eroded mask $\mathcal{M}_{e}$, the foreground likelihood should be one (left side of (\ref{eq:maskprior})).
This is motivated by the heuristic that pixels far away from the initial rendering should belong to the background, while pixels in the center of the initial rendering should belong to the foreground:

    \begin{equation}
    \mathcal{E}_{pri} = \big|(\mathcal{F} \circ \mathcal{M}_{e}) - \mathcal{M}_{e}\big|^2 + \big|\mathcal{F} \circ (1 - \mathcal{M}_{d})\big|^2,
    \label{eq:maskprior}
    \end{equation}
with $\circ$ being the Hadamard product.

\begin{figure*}[ht!]
\subcaptionbox{Illustration of the architecture of our approach. It consists of two sub-networks: a style encoder (left) and an animation network (middle).\label{fig:anim-arch-detailed}}{\includegraphics[width=1.0\columnwidth]{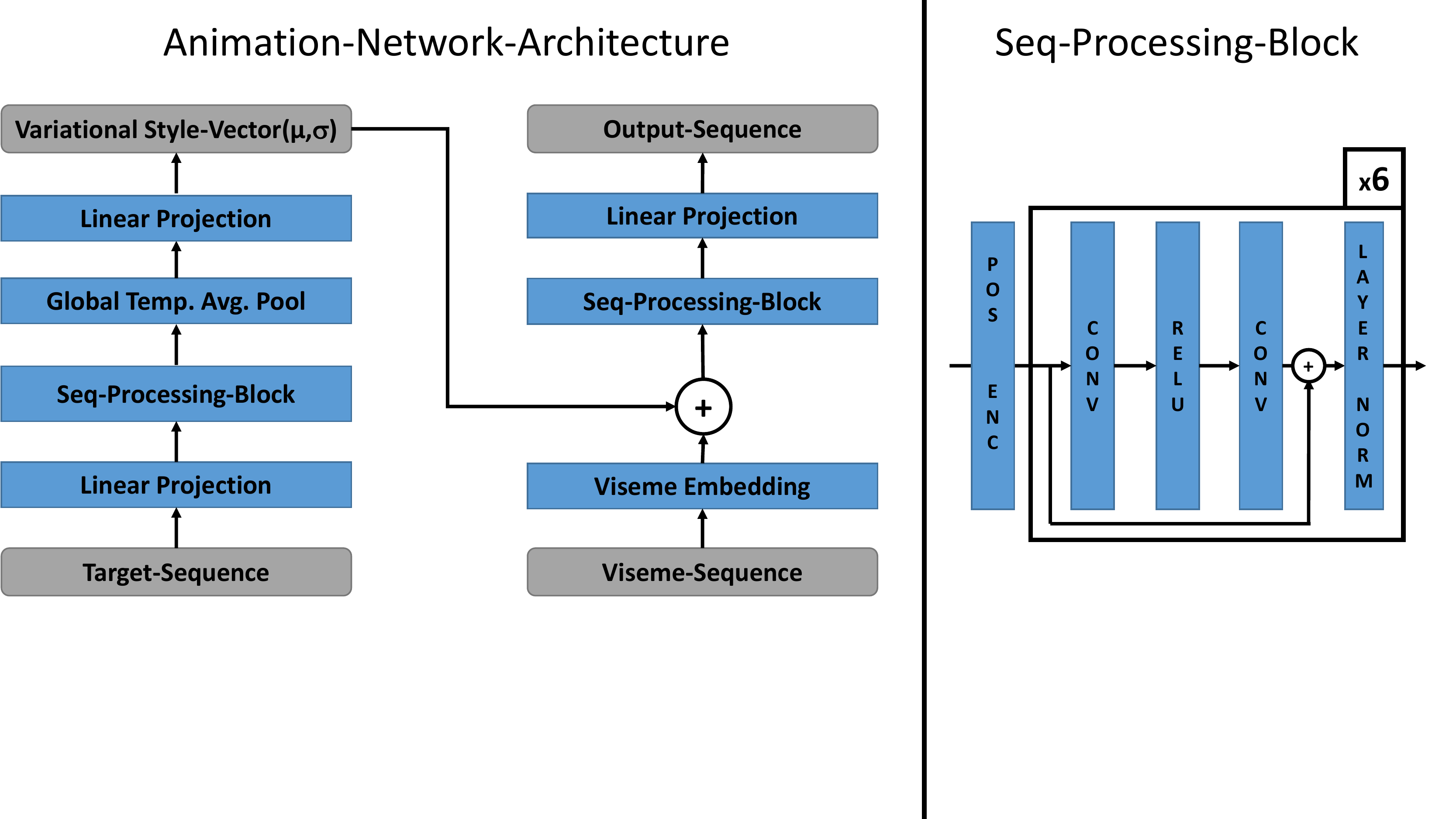}}\hfill%
\subcaptionbox{Illustration of the architecture of our variational sequence autoencoders for eye and pose parameters.\label{fig:vsae}}{\includegraphics[width=1.0\columnwidth]{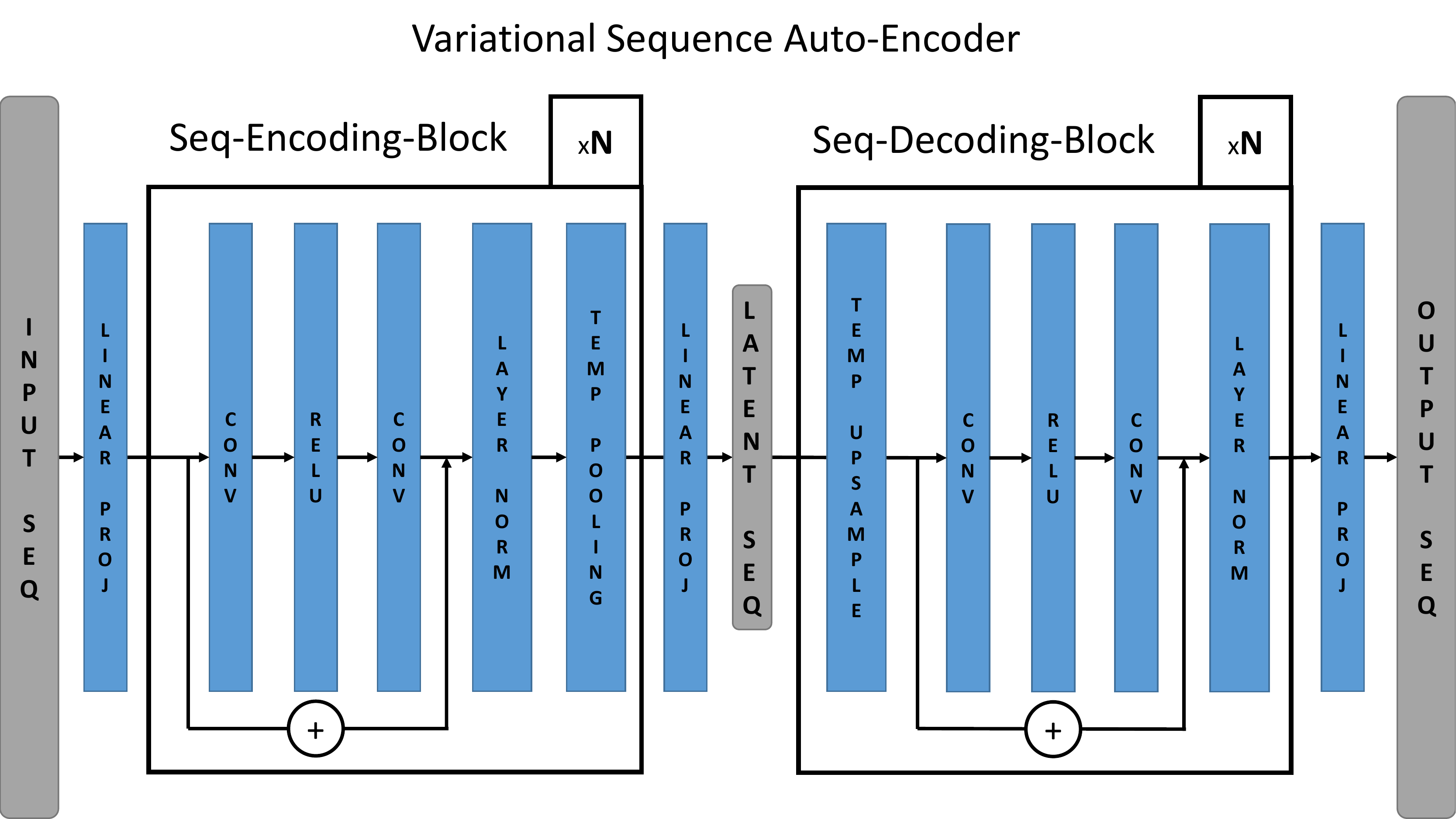}}%
\end{figure*}

Second, we implement an additional regularization term that encourages the network to generate a mostly binary foreground mask.
We introduce an additional $L_1$ loss that pulls the entries of $\mathcal{F}$ closer to zero if $\mathcal{F}_{i,j} < 0.5$ or to one if $\mathcal{F}_{i,j} > 0.5$, respectively.
During training, we start employing the mask regularization after 10 epochs by gradually increasing the regularization weight.
    \begin{equation}
    \mathcal{E}_{bin} =
    \begin{cases}
    \big|\mathcal{F}_{i,j}-1.0\big|,& \mathcal{F}_{i,j} > 0.5\\
    \big|\mathcal{F}_{i,j}\big|,      & \mathcal{F}_{i,j} < 0.5
	\end{cases}
    \label{eq:binarize}
    \end{equation}

Moreover, we control refinement by regularizing the corrective image $\mathcal{I}_{corr}$ with the $L_1$ norm.
    \begin{equation}
    \mathcal{E}_{reg} = \big|\mathcal{I}_{corr}\big|
    \label{eq:refinementreg}
    \end{equation}

\FloatBarrier
\section{Neural Face Animation}
\label{sec:animation}

This section presents our neural approach for realistic animation of our face model from text or speech.
For creating the animation model, we captured four multi-view sequences of our actress and asked her to present emotional monologues.
Additionally, we recorded the speech track with a small wireless microphone.
However, instead of training the animation model directly with the audio signal, we employ an existing forced alignment approach \cite{McAuliffe2017} that converts the speech track into timed phoneme sequences,
which are further transformed into viseme sequences that are in sync with the video data.
This is advantageous as the trained animation model can be used by more than just the captured person since visemes are speaker-agnostic.
Existing speech recognition software can be used to convert speech into viseme sequences, which allows for animation from audio, while tasks (e.g.~facial animation of sign language avatars) that benefit from a textual representation of speech can be easily implemented as well.

The input for the animation network consists of a sequence of viseme IDs where each viseme roughly approximates the visible mouth shape at a certain time frame.
Assuming 25 fps, the input sequence has $25 \cdot t$ entries where $t$ corresponds to the length of the animated sequence measured in seconds.
The output of our network is a 3D tensor of size $b\times (25 \cdot t)\times d$, where $b$ corresponds to the batch size and $d$ corresponds to the dimension of animation parameters.
This means that the duration of each expression/viseme is explicitly encoded in the input sequence by the number of repetitions.
Figure \ref{fig:anim-arch} shows a high-level architecture of our animation network.
It consists of three main structures, the core animation network that converts viseme IDs into animation parameters, a variational style encoder network that extracts a compact style vector from the target parameters signal, and two variational sequence autoencoder that act as animation priors for eye and pose parameters, which are not well correlated with the source viseme sequence.
\subsection{Variational Animation Prior}
\label{sec:animation-prior}
The main purpose of the variational sequence autoencoders (VSAE) for eye and pose parameters is to provide a statistical prior in order to minimize unnatural behavior.
Both VSAEs are based on a 1D convolutional architecture, which has two advantages: first, we benefit from better parallelization as well as faster training and second, it allows reducing the temporal resolution of the latent parameters via average pooling.
Reducing the temporal resolution in latent space forces the VSAE to learn more meaningful features, which yields smoother and more natural expression/motion sequences since each feature vector represents a short sequence of animation parameters.
Figure \ref{fig:vsae} shows the architecture of our VSAEs, where $N$ corresponds to the depth of the sequence autoencoder (i.e.~number of encoding/decoding blocks).
Via experiments, we found that setting $N=4$ for the pose sequence prior and $N=2$ for the eye sequence prior yields the best performance.
All convolutions have 128 filters, a kernel size of 3, and a padding of 1. The latent parameter sequences for eyes and pose have a feature dimension of 128 as well.
The objective function (\ref{eq:aniprior}) of the animation prior consists of the reconstruction loss and the Kullback-Leibler divergence between the distribution of latent sequence parameters $\mathcal{N}_{(\mu_z,\sigma_z)}$ and the standard normal distribution.
\begin{equation}
\mathcal{E}_{prior} = ||\textbf{X}-\hat{\textbf{X}}||^2 + \lambda_{prior} KL(\mathcal{N}_{(\mu_z,\sigma_z)}||\mathcal{N}_{(0,1)})
\label{eq:aniprior}
\end{equation}
\subsection{Style Aware Animation Network}
\label{sec:style-encoder}
The main purpose of the variational style encoder (figure \ref{fig:anim-arch-detailed}, left) is to provide additional information such that the animation network (figure \ref{fig:anim-arch-detailed}, middle) can disentangle content from speaking style and emotions.
This is important since we train our animation network with several speech sequences that include an arbitrary mix of different emotions and speaking styles.
Moreover, similar to a VAE, the style encoder learns a well-defined latent style space, which allows for simple editing and sampling of the speaking style vectors after training.
Again, the animation network and style encoder are based on a 1D convolutional architecture.
All convolutions have 128 filters, a kernel size of 9, padding of 4, and the latent style vector has 128 dimensions.
The objective function of the animation model~(\ref{eq:animodel}) consists of three data terms in order to minimize the prediction error of mouth expressions $\hat{\textbf{X}}_{mouth}$, latent eye expressions $\hat{\textbf{Z}}_{eyes}$ and latent pose $\hat{\textbf{Z}}_{pose}$.
The latent style vector is regularized via KL divergence such that the distribution of style vectors matches the standard normal distribution $\mathcal{N}_{(0,1)}$.
The latent pose $\textbf{Z}_{pose}$ and latent eye expressions $\textbf{Z}_{eyes}$ are computed by encoding the original pose and eye parameter sequences with the corresponding variational sequence autoencoders:

\begin{table}[]
\Large
\fontfamily{ptm}\selectfont
\resizebox{\columnwidth}{!}{%
\begin{tabular}{|lr|c|c|c|c|c|}
\hline
\rowcolor[HTML]{C0C0C0} 
\multicolumn{2}{|l|}{\cellcolor[HTML]{C0C0C0}\textbf{Visual Quality}} & \textbf{FOMM} & \textbf{FACIAL} & \textbf{NHA} & \textbf{4DFA} & \textbf{OURS}   \\ \hline
\textbf{L1}                    & \textbf{$\downarrow$}                & 0.0464        & 0.0469          & 0.0472       & 0.0345        & \textbf{0.0304} \\ \hline
\textbf{PSNR}                  & \textbf{$\uparrow$}                  & 25.133        & 23.495          & 24.219       & 25.111        & \textbf{26.127} \\ \hline
\textbf{SSIM}                  & \textbf{$\uparrow$}                  & 0.8155        & 0.8154          & 0.8038       & 0.8729        & \textbf{0.8872} \\ \hline
\textbf{LPIPS}                 & \textbf{$\downarrow$}                & 0.1581        & 0.0758          & 0.0937       & 0.1037        & \textbf{0.0510} \\ \hline
\end{tabular}%
}
\caption{Quantitative results of all tested rendering approaches based on traditional as well as learned error metrics.}
\label{fig:render-results2}
\end{table}

\begin{equation}
\begin{split}
\mathcal{E}_{anim} = ||\textbf{X}_{mouth}-\hat{\textbf{X}}_{mouth}||^2 + ||\textbf{Z}_{eyes}-\hat{\textbf{Z}}_{eyes}||^2\\
+||\textbf{Z}_{pose}-\hat{\textbf{Z}}_{pose}||^2 + \lambda_{style} KL(\mathcal{N}_{(\mu_{style},\sigma_{style})}||\mathcal{N}_{(0,1)})
\end{split}
\label{eq:animodel}
\end{equation}
All variables $\textbf{Z}$, $\textbf{X}$, $\hat{\textbf{Z}}$, and $\hat{\textbf{X}}$ represent sequences rather than single animation parameter/feature vectors.
Additionally, we train a small fully connected VAE (two hidden layers with 1024 neurons) that maps the original style vectors into a 2D style space.
This allows for visualizing the latent style space on a 2D grid and simplifies the task of finding suitable style vectors.
\subsection{Implementation and Training}
The neural face model uses 256-dimensional parameters for the mouth region, 256-dimensional parameters for the eye region, and 6-dimensional parameters for the head pose (orientation and position),
which results in a 518-dimensional animation parameter vector per frame.
That means for each frame the mouth expression, eye expression, as well as head, pose, and corresponding expression label~(viseme or idle) are known, while the emotion/style vector is learned automatically during training.
The full animation dataset consists of four short takes with a frame rate of 25 fps and a total length of 6500 frames (4.3 minutes). The first 500 frames of the third take are excluded from training and used for validation.
The animation network is trained with random sequence lengths between 2.5 seconds and 10 seconds.
Both animation priors are trained as independent auto-encoders, while the style-aware animation network and the style encoder are trained end-to-end.
All networks are trained simultaneously with a batch size of 32 and an initial learning rate of $0.001$ and $0.0005$, respectively.
For all networks, we use exponential learning rate scheduling with $\gamma=0.96$.
The animation networks are trained approximately for 25000 iterations with both regularization weights $\lambda_{prior}$ and $\lambda_{style}$ set to $1.0e^{-4}$.


\begin{table}[]
\Large
\fontfamily{ptm}\selectfont
\resizebox{\columnwidth}{!}{%
\begin{tabular}{|l|c|c|c|c|c|}
\hline
\rowcolor[HTML]{C0C0C0} 
\textbf{Visual Quality}  & \textbf{DAGAN} & \textbf{FACIAL} & \textbf{4DFA} & \textbf{OURS} & \textbf{GT} \\ \hline
\textbf{Mean}            & 2.5            & 2.4             & 3.6           & \textbf{4.0}  & 4.6         \\ \hline
\textbf{Median}          & 2.0            & 2.0             & \textbf{4.0}  & \textbf{4.0}  & 5.0         \\ \hline
\textbf{Very High, High} & 19\%           & 20\%            & 52\%          & \textbf{75\%} & 93\%        \\ \hline
\end{tabular}%
}
\caption{User ratings for the reconstructed videos, on a Likert scale (1 to 5). The first row shows the mean opinion score, the second row shows the median rating over all users, and the bottom row shows the percentage of how many participants gave a high (4) or very high rating(5).}
\label{fig:render-results4}
\end{table}

\section{Experimental Results and Discussion}
\label{sec:results}
This section presents our quantitative evaluation, our user study, and visual results generated with the proposed method as still images, while an accompanying video demonstrating dynamic effects can be found in the supplementary material.
For our experiments, we captured actors with synchronized and calibrated multi-view camera rigs consisting of three, five, or eight cameras that were equally placed around them at eye level.
The angle between neighboring cameras is always $45^\circ$, which yields a coverage of $360^\circ$ for the 8-camera rig, $225^\circ$ for five cameras, and $90^\circ$ for the 3-camera setup.
We captured different facial expressions as well as speech and asked them to present single words and short sentences in English.
The effective capture resolution for the head is approximately 520x360 pixels.
All pre-processing steps, network training, and experiments have been carried out on a regular desktop computer with 64GB Ram, 2.6 GHz CPU (14 cores with hyper-threading), and one GeForce RTX3090 graphics card.
For the creation of the personalized head models, we obtain a single 3D reconstruction of the person's head and adapt the geometry of an existing human body template (SMPL) \cite{SMPL2015}.
Personalized facial expression models are generated by transferring blend-shapes from an existing multi-person expression model (Facewarehouse)~\cite{Cao2014}.
We register them by optimizing similarity as well as identity parameters of the expression model.
With both models in correspondence, we obtain personalized expression shapes by sampling the 3D coordinates for all expression shapes for each face vertex of our head template.
For reasons of simplicity blend-shapes are transferred as differential shapes~(i.e.~the difference between neutral expression and corresponding expression shape).
All transferred blend-shapes are then compressed into a set of six expression shapes using PCA.
After generating a personalized head model, we use the facial performance capture pipeline described in section \ref{sec:capture} to obtain dynamic head meshes as well as dynamic head textures, which are used to train our hybrid generative head models as well as rendering models as described in section \ref{sec:capture} and section \ref{sec:rendering}.

\subsection{Evaluation of Modeling and Rendering}
\begin{table*}[]
\fontfamily{ptm}\selectfont
\resizebox{\textwidth}{!}{%
\begin{tabular}{|l|cccccc|}
\hline
\rowcolor[HTML]{C0C0C0} 
\textbf{VISUAL-QUALITY}                                                           & \multicolumn{1}{l|}{\cellcolor[HTML]{C0C0C0}\textbf{WAV-TO-LIP}} & \multicolumn{1}{c|}{\cellcolor[HTML]{C0C0C0}\textbf{MAKE-IT-TALK}} & \multicolumn{1}{c|}{\cellcolor[HTML]{C0C0C0}\textbf{FACIAL}} & \multicolumn{1}{c|}{\cellcolor[HTML]{C0C0C0}\textbf{AD-NERF}} & \multicolumn{1}{c|}{\cellcolor[HTML]{C0C0C0}\textbf{OURS}} & \textbf{GT} \\ \hline
\textbf{Mean}                                                                     & \multicolumn{1}{c|}{1.9}                                         & \multicolumn{1}{c|}{2.3}                                           & \multicolumn{1}{c|}{1.7}                                     & \multicolumn{1}{c|}{2.4}                                      & \multicolumn{1}{c|}{\textbf{3.8}}                          & 4.4         \\ \hline
\textbf{Median}                                                                   & \multicolumn{1}{c|}{2.0}                                         & \multicolumn{1}{c|}{2.0}                                           & \multicolumn{1}{c|}{2.0}                                     & \multicolumn{1}{c|}{2.0}                                      & \multicolumn{1}{c|}{\textbf{4.0}}                          & 5.0         \\ \hline
\textbf{Very-High, High}                                                          & \multicolumn{1}{c|}{9\%}                                         & \multicolumn{1}{c|}{13\%}                                          & \multicolumn{1}{c|}{5\%}                                     & \multicolumn{1}{c|}{18\%}                                     & \multicolumn{1}{c|}{\textbf{64\%}}                         & 89\%        \\ \hline
\rowcolor[HTML]{C0C0C0} 
\textbf{AUDIO-SYNC}                                                               & \textbf{}                                                        &                                                                    &                                                              &                                                               & \textbf{}                                                  & \textbf{}   \\ \hline
\textbf{Mean}                                                                     & \multicolumn{1}{c|}{2.4}                                         & \multicolumn{1}{c|}{2.2}                                           & \multicolumn{1}{c|}{2.1}                                     & \multicolumn{1}{c|}{1.7}                                      & \multicolumn{1}{c|}{\textbf{3.6}}                          & 4.5         \\ \hline
\textbf{Median}                                                                   & \multicolumn{1}{c|}{2.0}                                         & \multicolumn{1}{c|}{2.0}                                           & \multicolumn{1}{c|}{2.0}                                     & \multicolumn{1}{c|}{1.0}                                      & \multicolumn{1}{c|}{\textbf{4.0}}                          & 5.0         \\ \hline
\textbf{Very-High, High}                                                          & \multicolumn{1}{c|}{20\%}                                        & \multicolumn{1}{c|}{11\%}                                          & \multicolumn{1}{c|}{12\%}                                    & \multicolumn{1}{c|}{4\%}                                      & \multicolumn{1}{c|}{\textbf{56\%}}                         & 93\%        \\ \hline
\rowcolor[HTML]{C0C0C0} 
\textbf{\begin{tabular}[c]{@{}l@{}}FACIAL-EXPRESSION,\\ HEAD-MOTION\end{tabular}} &                                                                  &                                                                    &                                                              &                                                               & \textbf{}                                                  &             \\ \hline
\textbf{Mean}                                                                     & \multicolumn{1}{c|}{1.5}                                         & \multicolumn{1}{c|}{2.3}                                           & \multicolumn{1}{c|}{2.3}                                     & \multicolumn{1}{c|}{2.3}                                      & \multicolumn{1}{c|}{\textbf{3.6}}                          & 4.4         \\ \hline
\textbf{Median}                                                                   & \multicolumn{1}{c|}{1.0}                                         & \multicolumn{1}{c|}{2.0}                                           & \multicolumn{1}{c|}{2.0}                                     & \multicolumn{1}{c|}{2.0}                                      & \multicolumn{1}{c|}{\textbf{4.0}}                          & 5.0         \\ \hline
\textbf{Very-High, High}                                                          & \multicolumn{1}{c|}{3\%}                                         & \multicolumn{1}{c|}{8\%}                                           & \multicolumn{1}{c|}{17\%}                                    & \multicolumn{1}{c|}{14\%}                                     & \multicolumn{1}{c|}{\textbf{55\%}}                         & 89\%        \\ \hline
\end{tabular}%
}
\caption{User ratings for animated videos, on a Likert scale (from 1 up to 5). participants were asked to rate visual quality, synchronicity between mouth and voice, and the realism of facial expressions. For each approach, we compute the mean and median rating as well as the percentage of people who rated high or very high.}
\label{fig:animation-results-study}
\end{table*}
To assess the quality of the proposed hybrid modeling and rendering approach, we compare rendering results from our proposed approach and four existing methods \cite{Gafni2021, Siarohin2019, Zhang2021, Grassal2022} based on a challenging publicly available dataset used in the paper of Gafni et al.~\cite{Gafni2021}.
For the reason of fairness, we generated the results of all approaches based on monocular data.
While this limits the ability of our approach to modify the captured head-pose, it still allows for performing self-reenactment based on a monocular validation sequence.
Since we want to focus on the quality of the face and the head, we mask out the background and upper body.
Figure \ref{fig:render-results1} shows typical problems of the reference approaches: image-based methods like FOMM~\cite{Siarohin2019} work well as long as the head pose remains the same, however as soon as the actor looks in a different direction obvious rendering artifacts appear.
The second approach (FACIAL~\cite{Zhang2021}) is based on a pre-trained morphable head model. Since it is not able to capture facial expressions accurately enough, mouth and eye expressions can heavily deviate from the ground truth.
While the recent approach of Grassal et al.~\cite{Grassal2022} (NHA) captures the personal head geometry and appearance very well, glasses seem to pose a problem for this approach as the last row shows.
The NeRF-based approach of Gafni et al.~\cite{Gafni2021} (4DFA) works well on the validation sequence, however, the careful comparison shows also inaccurate eye expressions and a lack of details on skin and hair as well as a lower quality based on the evaluated metrics in table \ref{fig:render-results2}.

Figure \ref{fig:render-results3} shows a more detailed comparison between 4DFA and our approach.
As both methods learn to separate fore- and background during training, we also compare the generated foreground alpha masks.
The most obvious difference is that our masks are much clearer, but on a closer look, one can also see that our approach is able to reproduce more details on the skin and hair.
Table \ref{fig:render-results2} shows the evaluation of all rendering approaches with different error metrics such as $L_1$, PSNR, SSIM~\cite{Wang2004} and LPIPS (Learned Perceptual Image Patch Similarity)~\cite{Zhang2018}.
Especially the two perceptual error metrics (SSIM and LPIPS) show interesting results:
4DFA achieves only good structural similarity (0.8729), while FACIAL has a lower error according to LPIPS (0.0758).
In contrast, our approach performs best according to SSIM (0.8872) and LPIPS (0.0510).
A typical problem when evaluating dynamic visual content is that perceived image/video quality is difficult to measure with traditional error metrics.
Therefore, we conducted a user study where 70 participants (non-expert users, between 18 and 60 years old) were asked to rate the quality of the reconstructed head videos as well as the ground truth, see table \ref{fig:render-results4}.
For the user study, we compared our method against the approaches of Hong et al.~\cite{Hong2022} (DAGAN), Zhang et al.~\cite{Zhang2021} (FACIAL), Gafni et al.~\cite{Gafni2021} (4DFA), and the ground truth.
With each method, we generated four talking-head videos without sound and an approximate duration of four seconds.
Every participant was presented with all videos in random order and asked to rate the visual quality on a Likert scale (from 1 up to 5).
The ground truth achieved the highest mean opinion score (4.6), followed by our approach with a score of 4.0 and 4DFA with a score of 3.6.
Videos produced with DAGAN and FACIAL received lower ratings (2.5 and 2.4) on average,
which was probably caused by temporal artifacts/flickering and unnatural deformations of the face due to strong head movement (DAGAN).
Summarized, quantitative evaluation as well as the user study show that our modeling and rendering approach can generate high-quality renderings and outperforms all reference approaches.
\subsection{Evaluation of our Neural Animation Approach}
\begin{figure*}[t]
\includegraphics[trim=0mm 0mm 0mm 0mm,clip,width=\textwidth]{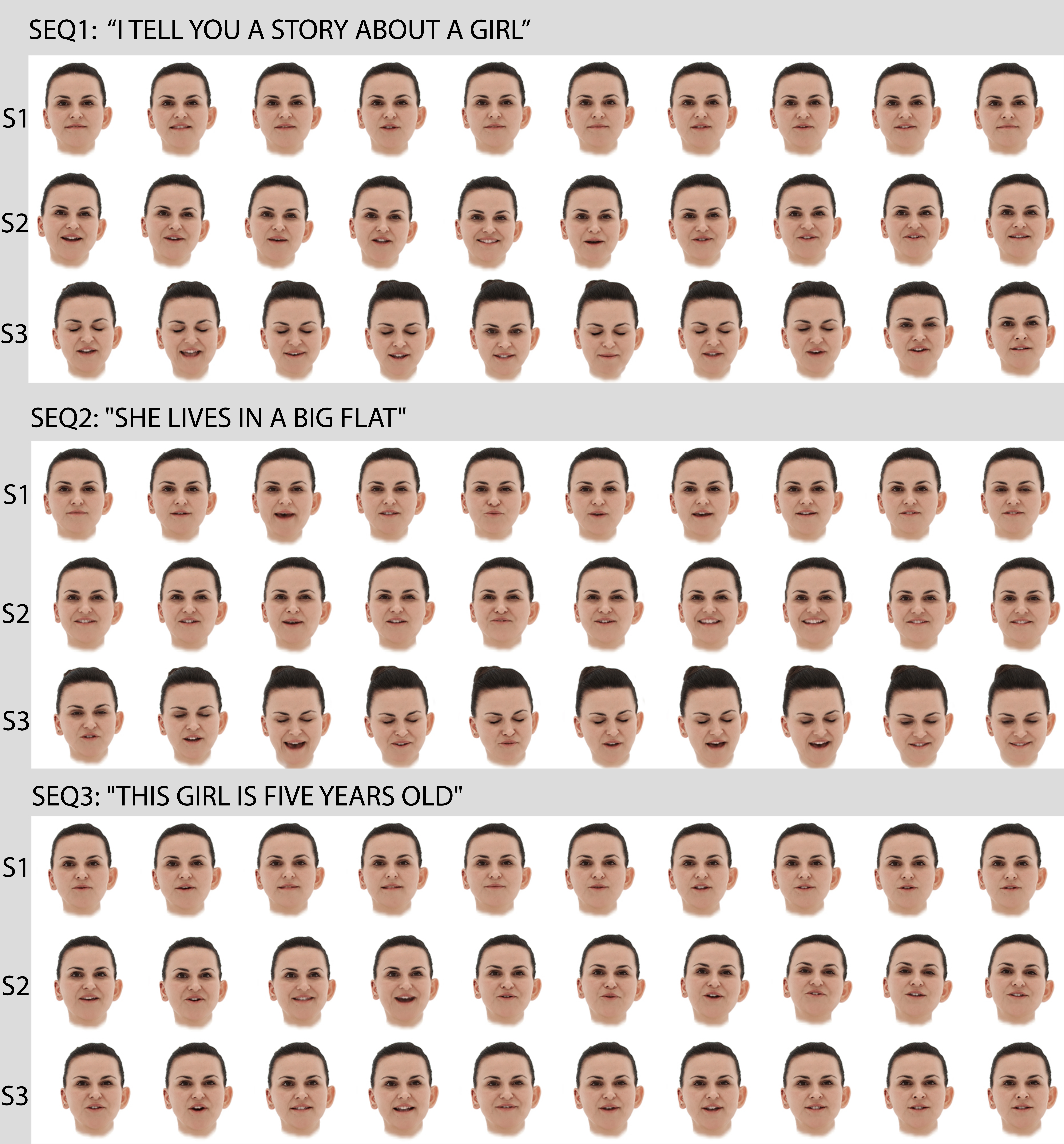}
\protect\caption{This figure shows three animated sequences based on our captured audio-visual dataset. Each sentence was animated with three different style vectors that could be described as 'neutral' (S1), 'happy' (S2), and 'angry/frustrated' (S3). }
\label{fig:animation-results}
\end{figure*}

\begin{table*}[]
\fontfamily{ptm}\selectfont
\resizebox{\textwidth}{!}{%
\begin{tabular}{|ll|l|l|l|l|l|}
\hline
\rowcolor[HTML]{C0C0C0} 
\multicolumn{2}{|l|}{\cellcolor[HTML]{C0C0C0}\textbf{AV-Synch}} & \textbf{WAV-TO-LIP} & \textbf{MAKE-IT-TALK} & \textbf{FACIAL} & \textbf{AD-NERF} & \textbf{OURS} \\ \hline
\textbf{LMD}                         & \textbf{$\downarrow$}    & 11.8                & 11.3                  & 14.2            & 13.7             & \textbf{6.1}  \\ \hline
\textbf{Abs. Offset (SynchNet)}           & \textbf{$\downarrow$}    & 3                  & 2                    & 2               & \textbf{1}      & \textbf{1}   \\ \hline
\textbf{AV-Confidence (SynchNet)}    & \textbf{$\uparrow$}      & \textbf{6.6}        & 3.7                   & 2               & 2.2              & 4.2           \\ \hline
\end{tabular}%
}
\caption{Quantitative measures for the synchronicity between speech and mouth movements. LMD corresponds to the mean-squared distance between landmarks of the animated face and the captured reference video,~\cite{Chen2018}. Additionally, we compare the predicted frame-offset and the audio-visual confidence predicted by SynchNet~\cite{Chung2017}.}
\label{fig:av-synch}
\end{table*}

The evaluation of the proposed animation approach is based on our audio-visual multi-view dataset.
Figure \ref{fig:animation-results} shows three short animated speech sequences that were produced with our method. Each sequence was rendered based on three manually chosen style vectors that represent 'neutral' speech style (S1), angry/frustrated speech (S2), and happy/delighted speech (S3).
This demonstrates that our system can disentangle content and speaking style in an unsupervised manner, which allows for synthesizing of emotional speech sequences and gives manual control over style/emotion.
Moreover, we show that the learned style space allows consistently animating the head model, as the same style vectors yield similar emotions/speaking styles also with different input sequences.
To evaluate the quality of the proposed animation approach, we mainly rely on the user study as this is the most meaningful way of measuring the perceived quality of synthesized talking head videos.
We compare our animation results against the ground-truth and synthesized talking head videos generated with four recent speech-driven animation methods (Wav-To-Lip~\cite{Prajwal2020}, Make-It-Talk~\cite{Zhou2020}, Facial~\cite{Zhang2021}, and AD-Nerf~\cite{Guo2021}).
In an online questionnaire~(implemented with the SoSciSurvey online survey platform~\cite{SOSI}), we asked 70 English-speaking participants (non-expert users, between 18 and 60 years old) to rate three important characteristics of each synthesized video: the rendering quality, the synchronicity between mouth and voice, and the realism of facial expressions.
Based on our captured emotional-speech data, we generated three short speech sequences (approx. 4 seconds long) with each reference method.
All video stimuli (including the corresponding captured speech videos of the actress) were presented in randomized order and at the same resolution (600x600 pixels).
In order to improve quality of the collected data, each participant was first presented a training task (incl. instructions) based on the synthetic talking head video of a different actor.
Video stimuli could be played only once. To reduce distraction, answer options appeared after the video finished.
Table \ref{fig:animation-results-study} shows that our approach achieves the highest ratings (except ground truth) in all three categories.
In addition, we conduct a quantitative evaluation of the audio-visual synchronicity (table \ref{fig:av-synch}) based on the mean-squared landmark distance (LMD) proposed by Chen et al.~\cite{Chen2018} as well as the predicted frame-offset and the audio-visual confidence measure proposed by Chung et al.~\cite{Chung2017}.
As our reference videos contain considerable head motion, we compute the LMD on registered sets of mouth landmarks (i.e. after compensating for 2D-translation, 2D-rotation and uniform-scale).
Assuming perfect audio-visual synchronicity in the captured reference videos, a high LMD value indicates a high deviation between synthesized and captured mouth movements, while an LMD close to zero indicates almost identical mouth movements.
The method of Chung et al.~\cite{Chung2017} predicts a frame offset in order to align a video with its audio track.
Additionally, a confidence value (unnormalized) measures the correspondence between video and audio track.
A high confidence (e.g. 6) indicates well matching speech and mouth movements, while a confidence close to zero indicates that mouth movements are uncorrelated with the audio.
Again our approach outperforms the reference methods achieving the lowest LMD error (6.1) and the lowest predicted absolute frame-offset of 1 with a confidence of 4.2.
Only WAV-TO-LIP~\cite{Prajwal2020} achieved higher confidence (6.6) but also a higher offset of 3 frames.

It has to be emphasized that the training videos are challenging as our actress used a lot of body language to convey emotions, which resulted in large movements and very strong facial expressions.
In the supplemental video, we show samples of the training data, rendering and animation results of reference approaches, ablation studies to visualize the learned style space and the impact of the learned animation prior models as well as the variational style encoder.
\FloatBarrier
\subsection{Limitations and Future Work}
A limitation of the proposed approach is the missing flexibility in terms of lighting conditions during rendering as the studio light is baked into the textures.
For future versions of our system we plan to further extend our approach such that the light situation can be adapted at render time according to the target scene.
Regarding animation, our system is limited to emotions and speaking styles that are presented by the captured actors.
Training a multi-person animation model that disentangles animation style from actor identity would further increase the flexibility of our approach as this could allow extrapolating/transferring animation styles between actors.

\section{Conclusions}
\label{sec:conclusion}
We present a new method for the creation and animation of photo-realistic 3D human head models.
Our hybrid head model combines the advantages of lightweight and robust model-based representations and the realism of neural rendering techniques, which allows for animation, facial expression editing as well as realistic rendering in real time.
The self-supervised architecture allows for convenient training (without pre-computed segmentation masks) and supports multi-view data.
Moreover, automatically learned foreground/background segmentation masks allow for simple integration of rendered head images with new backgrounds as well as with 3D/VR applications.
Based on the neural head representation, we introduce a novel style-aware facial animation approach that uses a sequence-to-sequence translation architecture.
In contrast to previous methods, our animation network can be successfully trained with real acting performances that contain strong emotions (unannotated) and as well as large head movements.
To synthesize more realistic videos, we simultaneously train variational sequence auto-encoders for eye expression and head pose, which are typically not well correlated with raw speech content (text, phonemes, visemes)
The VSAEs act as a statistical animation prior, which results in smoother and more natural expression/motion sequences.
An additional style encoder helps to capture natural variations of speech as well as emotions via low-dimensional latent style vectors that can be used after training to control the animated speech style.
We show that our approach successfully disentangles content and style via a consistent latent style space and evaluate our animation method as well as the proposed hybrid head model quantitatively and qualitatively on challenging datasets.
Moreover, we conducted a study showing that the proposed rendering as well as the animation approach receive the highest ratings in terms of perceived quality from 70 participants.
\FloatBarrier

\printcredits

\section*{Acknowledgements}
This work has partly been funded by the European Union’s Horizon 2020 research and innovation programme under grant agreement No 952147 (Invictus),
by the German Federal Ministry of Education and Research (Voluprof, grant no.~16SV8705),
by the German Federal Ministry for Economic Affairs and Climate Action (ToHyVe, grant no.~01MT22002A)
as well as the Fraunhofer Society in the Max Planck-Fraunhofer collaboration project NeuroHum.

\bibliographystyle{plain}

\bibliography{paper-graphical-models.bib}

\end{document}